\documentclass[10pt,twocolumn,letterpaper]{article}
\usepackage[table,xcdraw]{xcolor}
\usepackage{titling}
\usepackage{booktabs}
\usepackage{subcaption}
\usepackage{cvpr}
\usepackage{times}
\usepackage{epsfig}
\usepackage{graphicx}
\usepackage{amsmath}
\usepackage{amssymb}
\usepackage{url}
\usepackage{textcomp}
\usepackage{algorithm}
\usepackage[noend]{algpseudocode}

\usepackage{breakurl}
\usepackage[pagebackref=true,breaklinks=true,letterpaper=true,colorlinks,bookmarks=false]{hyperref}
\makeatletter
\def\BState{\State\hskip-\ALG@thistlm}
\makeatother
\cvprfinalcopy % *** Uncomment this line for the final submission

\def\cvprPaperID{1} % *** Enter the CVPR Paper ID here

\ifcvprfinal\fi
\begin{document}

%%%%%%%%% TITLE
\title{A Realistic Dataset and Baseline Temporal Model for Early Drowsiness Detection}

\author{Reza Ghoddoosian \hspace{2cm} Marnim Galib \hspace{2cm} Vassilis Athitsos \\
Vision-Learning-Mining Lab, University of Texas at Arlington\\
{\tt\small \{reza.ghoddoosian, marnim.galib\}@mavs.uta.edu, athitsos@uta.edu
}
}

\maketitle
\thispagestyle{empty}

%%%%%%%%% ABSTRACT
\begin{abstract}
Drowsiness can put lives of many drivers and workers in danger. It is important to design practical and easy-to-deploy real-world systems to detect the onset of drowsiness. In this paper, we address early drowsiness detection, which can provide early alerts and offer subjects ample time to react. We present a large and public real-life dataset\footnote {Available on: sites.google.com/view/utarldd/home} of 60 subjects, with video segments labeled as alert, low vigilant, or drowsy. This dataset consists of around 30 hours of video, with contents ranging from subtle signs of drowsiness to more obvious ones. We also benchmark a temporal model\footnote {Code available on: https://github.com/rezaghoddoosian} for our dataset, which has low computational and storage demands. The core of our proposed method is a Hierarchical Multiscale Long Short-Term Memory (HM-LSTM) network, that is fed by detected blink features in sequence. Our experiments demonstrate the relationship between the sequential blink features and drowsiness. In the experimental results, our baseline method produces higher accuracy than human judgment.
\end{abstract}

%%%%%%%%% BODY TEXT
\section{Introduction}
Drowsiness detection is an important problem. Successful solutions have applications in domains such as driving and workplace. For example, in driving, National Highway Traffic Safety Administration in the US estimates that 100,000 police-reported crashes are the direct result of driver fatigue each year. This results in an estimated 1,550 deaths, 71,000 injuries, and \$12.5 billion in monetary losses~\cite{16}.
To put this into perspective, an estimated 1 in 25 adult drivers report having fallen asleep while driving in the previous 30 days~\cite{18,17}.
In addition, studies show that, when driving for a long period of time, drivers lose their self-judgment on how drowsy they are~\cite{14}, and this can be one of the reasons that many accidents occur close to the destination. Research has also shown that sleepiness can affect workers' ability to perform their work safely and efficiently~\cite{workplace,fatigue}. All these troubling facts motivate the need for an economical solution that can detect drowsiness in early stages. It is commonly agreed \cite{8,9,11} that there are three types of sources of information in drowsiness detection:
Performance measurements, physiological measurements, and behavioral measurements.

\begin{figure}[t]
\begin{center}

   \includegraphics[width=0.7\linewidth]{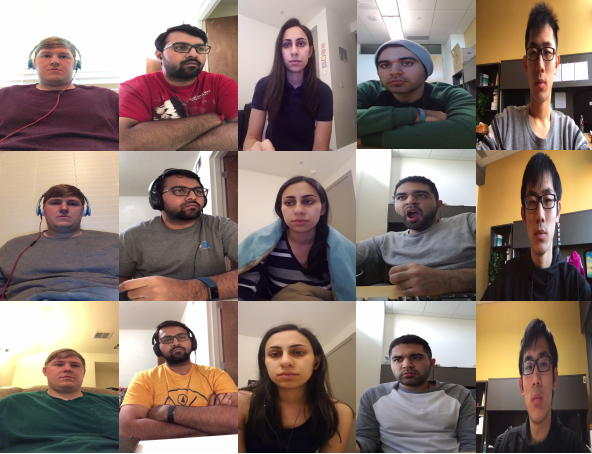}
\end{center}
   \caption{Sample frames from the RLDD dataset in the alert (first row), low vigilant (second row) and drowsy (third row) states.}
\label{sample}
\end{figure}

For instance, in the driving domain, performance measurements focus on steering wheel movements, driving speed, brake patterns, and lane deviations. An example is the Attention Assist system by  Mercedes Benz~\cite{20}. As practical as these methods can be, such technologies are oftentimes reserved for high-end models, as they are too expensive to be accessible to the average consumer.
Performance measurements at workplace can be obtained by testing workers' reaction time and short-term memory~\cite{fatigue}. Physiological measurements such as heart rate, electrocardiogram (ECG), electromyogram (EMG), electroencephalogram (EEG)~\cite{22,15} and electrooculogram (EOG)~\cite{15} can be used to monitor drowsiness. However, such methods are intrusive and not practical to use in the car or workspace despite their high accuracy. Wearable hats have been proposed as an alternative for such measurements ~\cite{21}, but they are also not practical to use for long hours.

Behavioral measurements are obtained from facial movements and expressions using non-intrusive sensors like cameras. In Johns's work~\cite{5}, blinking parameters are measured by light-emitting diodes. However, this method is sensitive to occlusions, where some object such as a hand is placed between the light emitting diode and the eyes. 

Phone cameras are an accessible and cheap alternative to the aforementioned methods. One of the goals of this paper is to introduce and investigate an end-to-end processing pipeline that uses input from phone cameras to detect both subtle and more clearly expressed signs of drowsiness in real time. This pipeline is computationally cheap so that it could ultimately be implemented as a cell phone application available for the general public. 

Previous work in this field mostly focused on detecting extreme drowsiness with explicit signs such as yawning, nodding off and prolonged eye closure~\cite{7,9,13}. However, for drivers and workers, such explicit signs may not appear until only moments before an accident. Thus, there is significant value in detecting drowsiness at an early stage, to provide more time for appropriate responses. The proposed dataset represents subtle facial signs of drowsiness as well as the more explicit and easily observable signs, and thus it is an appropriate dataset for evaluating early drowsiness detection methods. 

Our data consists of around 30 hours of RGB videos, recorded in indoor real-life environments by various cell phone/web cameras. The frame rates are below 30 fps, which makes drowsiness detection more challenging, as blinks are not observed as clearly as in high frame-rate videos.
The videos in the dataset are labeled using three class labels: alertness, low vigilance, and drowsiness (Fig.\ref{sample}). The videos have been obtained from 60 participants. The need for research in early drowsiness detection is further illustrated by experiments we have conducted, where we asked twenty individuals to classify videos from our dataset into the three predefined classes. The average accuracy of the human observers was under 60\%. This low accuracy indicates the challenging nature of the early drowsiness detection problem.

In addition to contributing a large and public realistic drowsiness dataset, we also implement a baseline method and include quantitative results from that method in the experiments. The proposed method leverages the temporal information of the video using a Hierarchical Multiscale LSTM (HM-LSTM) network~\cite{23} and voting, to model the relationship between blinking and state of alertness. The proposed baseline method produces higher accuracy than human judgment in our experimental results. 

Previous work on drowsiness detection produced results on datasets that were either private \cite{4} or acted \cite{7,9}. By ``acted'' we mean data where subjects were instructed to simulate drowsiness, compared to ``realistic'' data, such as ours, where subjects were indeed drowsy in the corresponding videos. The lack of large, public, and realistic datasets has been pointed out by researchers in the field~\cite{11,7,9}. 

Our work is motivated to some extent by the driving domain (i.e., camera angle and distance in our dataset, and the calibration period in our method as explained in Sec. 4.2). However, our dataset has not been obtained from driving and it does not capture some important aspects of driving such as night lighting and camera vibration due to car motion. Given these aspects of our dataset, we do not claim that our dataset and results represent driving conditions. At the same time, the data and the proposed baseline method can be useful for researchers targeting other applications of drowsiness detection, for example in workplace environments.

The proposed dataset offers significant advantages over existing public datasets for drowsiness detection, regardless of whether those existing datasets have been motivated by the driving domain or not: (a) it is the largest to date public drowsiness detection dataset,
 (b) the drowsiness samples are real drowsiness as opposed to acted drowsiness in~\cite{25}, and
(c) the data were obtained using different cameras. Each subject recorded themselves using their cell phone or web camera, in an indoor real-life environment of their choice. This is in contrast to existing datasets~\cite{25,22} where recordings were made in a lab setting, with the same background, camera model, and camera position.

Other contributions of this paper can be summarized as follows: (a) introducing, as a baseline method, an end-to-end real time drowsiness detection pipeline based on low frame rates resulting in a higher accuracy than that of human observers, and
(b) combining blinking features with Hierarchical Multiscale Recurrent Neural Networks to tackle drowsiness detection using subtle cues. These cues, which can be easily missed by human observers, are useful for detecting the onset of drowsiness at an early stage, before it reaches dangerous levels.

\begin{figure*}
\begin{flushleft}
 \includegraphics[width=1.0\linewidth,height=2.1in]{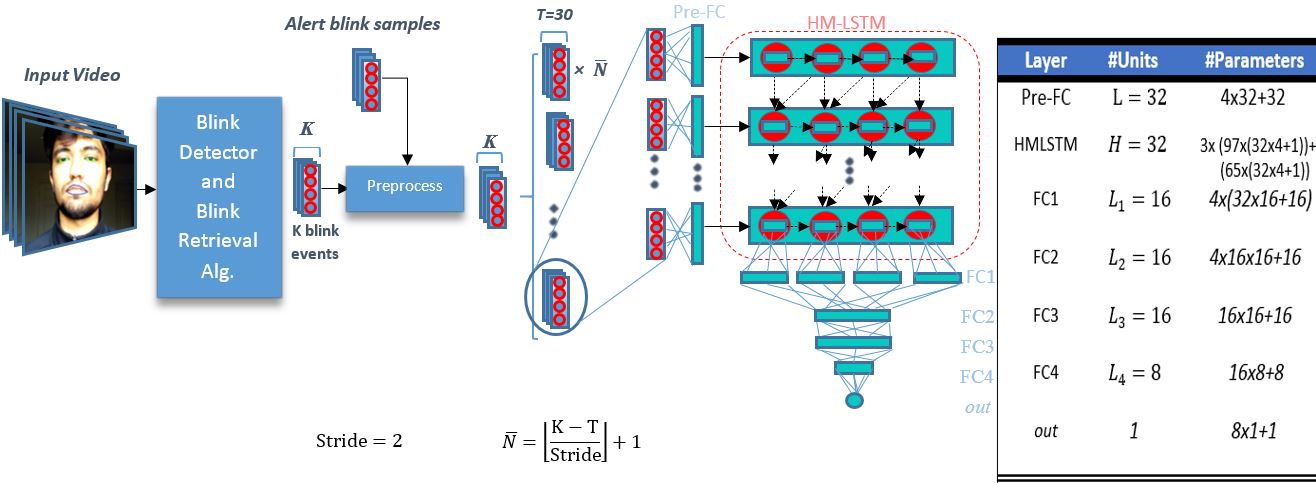}
\end{flushleft}
   \caption{The model design and configuration.}
\label{architecture}
\end{figure*}

\section{Related Work}

Drowsiness Detection has been studied over several years. In the rest of this section, a review of the available datasets and existing methods will be provided.
\subsection{Datasets}

As pointed out above, there are numerous works in drowsiness detection, but none of them uses a dataset that is both  public and realistic. As a result, it is difficult to compare prior methods to each other and to decide what the state of the art is in this area. Several existing methods~\cite{5,1,2,8,10} were evaluated on a small number of subjects without sharing the videos. In some cases~\cite{6,9} the subjects were instructed to act drowsy, as opposed to obtaining data from subjects who were really drowsy.

Some datasets~\cite{ME1,ME2,ME3} have been created for short and general micro expression detection which are not applicable specifically for drowsiness detection. The NTHU-driver drowsiness detection dataset is a public dataset which contains IR videos of 36 participants while they simulate driving~\cite{25}. However, it is based on subjects pretending to be drowsy, and it is an open question whether and to what extent videos of pretended drowsiness are useful training data for detecting real drowsiness, especially at an early stage.

The DROZY dataset~\cite{22}, contains multiple types of drowsiness-related data including signals such as EEG, EOG and near-infrared (NIR) images. An advantage of the DROZY dataset is that drowsiness data are obtained by subjects who are really drowsy, as opposed to pretending to be drowsy. Compared to the DROZY dataset, our dataset has three advantages: First, we have a substantially larger number of subjects (60 as opposed to 14). Second, for each subject, we have data showing that subject in each of the three predefined alertness classes, whereas in the DROZY dataset some subjects are not recorded in all three states. Third, in DROZY all videos were captured using the same camera position and background, under controlled lab conditions, whereas in our dataset each subject used their own cell phone and a different background. Compared to DROZY, our dataset also has the important difference that it provides color video, whereas DROZY offers several other modalities, but only NIR video.

Last but not least, Friedrichs and Yang~\cite{4}, used 90 hours of real driving to train and evaluate their method, but their dataset is private and not available as a benchmark.

\subsection{Drowsiness Detection Methods}

Features in non-intrusive drowsiness detection by cameras are divided into handcrafted features or features learned automatically using CNNs. Regarding handcrafted features, the most informative facial region about drowsiness is the eyes, and commonly used features are usually related to blinking behavior. McIntire \etal~\cite{1} show how blink frequency and duration normally increase with fatigue by measuring the reaction time and using an eye tracker. Svensson \cite{15} has shown that the amplitude of blinks can also be an important factor. Friedrichs and Yang \cite{4} investigate many blinking features like eye opening velocity, average eye closure speed, blink duration, micro sleeps and energy of blinks as well as head movement information. They report a final classification rate of 82.5\% on their own private dataset, which is noticeably larger than the 65.2\% accuracy that we report in our experiments. However, all the features in \cite{4} are extracted using the Seeing Machines sensor \cite{27} that uses not only video information (with the frame rate of 60 fps) but also the speed of the car, GPS information and head movement signals to detect drowsiness. In contrast, in our work the data come from a cell phone/web camera.

Recent research examines the effectiveness of Deep Neural Networks for end-to-end feature extraction and drowsiness detection, as opposed to the works that use handcrafted features with conventional classifiers or regressors such as regression and discriminant analysis (LDA)~\cite{2}, or fitting a 2D Gaussian with thresholding~\cite{5}. The results of the mentioned studies were not validated based on a large or public dataset.

Park \etal~\cite{7} fine-tune three CNNs and apply an SVM to the combined features of those three networks to classify each frame into four classes of alert, yawning, nodding and drowsy with blinking. The model is trained on the NTHU drowsiness dataset that is based on pretended drowsiness, and tested on the evaluation portion of NTHU dataset which includes 20 videos of only four people, resulting in 73\% drowsiness detection accuracy. We should note that the accuracy we report in our experiment is 65.2\%, which is lower that the 73\% accuracy reported in~\cite{7}. However, the method of~\cite{7} was evaluated on pretended data, where the signs of drowsiness tend to be easily visible and even exaggerated. Also, the work of Park \etal does not consider pooling the temporal information in the videos and classifies each frame independently, thus it can only classify based on the clear signs of drowsiness.

Bhargava \etal~\cite{9} show how a distilled deep network can be of use for embedded systems. This is relevant to the baseline method proposed in this paper, which also aims for low computational requirements. The reported accuracy in~\cite{9} is 89\% using three classes (alert, yawning, drowsy), based on training on patches of eyes and lips. Similar to Park \etal's work, Bhargava \etal's network also classifies each frame independently, thus not using temporal features. The dataset they used is private, and based on acted drowsiness, so it is difficult to compare those results to the results reported in this paper.
%-------------------------------------------------------------------------
\section{The Real-Life Drowsiness Dataset (RLDD)}
\subsection{Overview}
The RLDD dataset was created for the task of multi-stage drowsiness detection, targeting not only extreme and easily visible cases, but also subtle cases of drowsiness. Detection of these subtle cases can be important for detecting drowsiness at an early stage, so as to activate drowsiness prevention mechanisms. Our RLDD dataset is the largest to date realistic drowsiness dataset.

The RLDD dataset consists of around 30 hours of RGB videos of 60 healthy participants. For each participant we obtained one video for each of three different classes: alertness, low vigilance, and drowsiness, for a total of 180 videos. Subjects were undergraduate or graduate students and staff members who took part voluntarily or upon receiving extra credit in a course. All participants were over 18 years old. There were 51 men and 9 women, from different ethnicities (10 Caucasian, 5 non-white Hispanic, 30  Indo-Aryan and Dravidian, 8 Middle Eastern, and 7 East Asian) and ages (from 20 to 59 years old with a mean of 25 and standard deviation of 6). The subjects wore glasses in 21 of the 180 videos, and had considerable facial hair in 72 out of the 180 videos. Videos were taken from roughly different angles in different real-life environments and backgrounds. Each video was self-recorded by the participant, using their cell phone or web camera. The frame rate was always less than 30 fps, which is representative of the frame rate expected of typical cameras used by the general population.

\subsection{Data Collection}
In this section we describe how we collected the videos for the RLDD dataset. Sixty healthy participants took part in the data collection. After signing the consent form, subjects were instructed to take three videos of themselves by their phone/web camera (of any model or type) in three different drowsiness states, based on the KSS table~\cite{29} (Table \ref{kss}), for around ten minutes each. The subjects were asked to upload the videos as well as their corresponding labels on an online portal provided via a link. Subjects were given ample time (20 days) to produce the three videos. Furthermore, they were given the freedom to record the videos at home or at the university, any time they felt alert, low vigilant or drowsy, while keeping the camera set up (angle and distance) roughly the same. All videos were recorded in such an angle that both eyes were visible, and the camera was placed within a distance of one arm length from the subject. These instructions were used to make the videos similar to videos that would be obtained in a car, by phone placed in a phone holder on the dash of the car while driving. The proposed set up was to lay the phone against the display of their laptop while they are watching or reading something on their computer. 
\begin{table}
\footnotesize
\centering
\begin{tabular}{l}
\rowcolor[HTML]{32CB00} 
1- Extremely alert                       \\
\rowcolor[HTML]{32CB00} 
2- Very alert                            \\
\rowcolor[HTML]{32CB00} 
3- Alert                                 \\
4- Rather alert                          \\
5- Neither alert nor sleepy              \\
\rowcolor[HTML]{3166FF} 
6- Some signs of sleepiness              \\
\rowcolor[HTML]{3166FF} 
7- Sleepy, no difficulty remaining awake \\
\rowcolor[HTML]{F8FF00} 
8- Sleepy, some effort to keep alert     \\
\rowcolor[HTML]{F8FF00} 
9- Extremely sleepy, fighting sleep     
\end{tabular}
\caption{KSS drowsiness scale}\label{kss}
\end{table}
After a participant uploaded the three videos, we watched the entire videos to verify their authenticity and to make sure that our instructions were followed. In case of any question, we contacted the participants and asked them to share more details about the situation under which they recorded each video. In some cases, we asked them to redo the recordings and if the videos were clearly not realistic (people faking drowsiness as opposed to being drowsy) or off the standard, we simply ignored those videos for quality reasons. The three classes were explained to the participants as follows:

1) \textbf{Alert}: One of the first three states highlighted in the KSS table in Table \ref{kss}. Subjects were told that being alert meant they were experiencing no signs of sleepiness.

2) \textbf{Low Vigilant}: As stated in level 6 and 7 of Table \ref{kss}, this state corresponds to subtle cases when some signs of sleepiness appear, or sleepiness is present but no effort to keep alert is required.

3) \textbf{Drowsy}: This state means that the subject needs to actively try to not fall asleep (level 8 and 9 in Table \ref{kss}).

\subsection{Content}
This dataset consists of 180 RGB videos. Each video is around ten minutes long, and is labeled as belonging to one of three classes: alert (labeled as 0), low vigilant (labeled as 5) and drowsy (labeled as 10). The labels were provided by the participants themselves, based on their predominant state while recording each video. Clearly there is a subjective element in deciding these labels, but we did not find a good way to remedy that problem, given the absence of any sensor that could provide an objective measure of alertness. This type of labeling takes into account and emphasizes the transition from alertness to drowsiness. Each set of videos was recorded by a personal cell phone or web camera resulting in various video resolutions and qualities. The 60 subjects were randomly divided into five folds of 12 participants, for the purpose of cross validation. The dataset has a total size of 111.3 Gigabytes.
\subsection{Human Judgment Baseline}\label{human}
We conducted a set of experiments to measure human judgment in multi-stage drowsiness detection. In these experiments, we asked four volunteers per fold (20 volunteers in total) to watch the unlabeled and muted videos in each fold and write down a real number between 0 to 10 estimating the drowsiness degree per video (see Table \ref{kss}). Before the experiment, volunteers (8 female and 12 male, 3 undergraduates and 17 graduate students) were shown some sample videos that illustrated the drowsiness scale. Then, they were left alone in a room to watch the videos (they were allowed to rewind back or fast forward the videos at will) and annotate them. In order to make sure that each judgment was independent of the other videos of the same person, volunteers were instructed to annotate one video of each subject before annotating a second video for any subject. Results of these experiments are demonstrated in section \ref{humanexp} and compared with the results of our baseline method. Observers (aged 26.1 $\pm$ 2.9 (mean $\pm$ SD)) were from computer science, psychology, nursing, social work and information systems majors.
\section{The Proposed Baseline Method}

In this section, we discuss the individual components of our proposed multi-stage drowsiness detection pipeline. The blink detection and blink feature extraction are described first. Then we discuss how we integrate a Hierarchical Multiscale LSTM module into our model, how we formulate drowsiness detection initially as a regression problem, and how we discretize the regression output to obtain a classification label per video segment. Finally, we discuss the voting process that is applied on top of classification results of all segments of a video.

\subsection{Blink Detection and Blink Feature Extraction}

The motivation behind using blink-related features such as blink duration, amplitude, and eye opening velocity, was to capture temporal patterns that  appear naturally in human eyes and might be overlooked by spatial feature detectors like CNNs (as it is the case for human vision shown in our experiments). We used dlib's pre-trained face detector based on a modification to the standard Histogram of Oriented Gradients + Linear SVM method for object detection~\cite{30}.

We improved the algorithm by Soukupov\'{a} and Cech~\cite{24} to detect eye blinks, using six facial landmarks per eye described in~\cite{31} to extract consecutive quick blinks that were initially missed in Soukupov\'{a} and Cech's work. Kazemi and Sullivan's~\cite{31} facial landmark detector is trained on an ``in-the-wild dataset'', thus it is more robust to varying illumination, various facial expressions, and moderate non-frontal head rotations, compared to correlation matching with eye templates or a heuristic horizontal or vertical image intensity projection~\cite{24}.
In our experiments, we noticed that the approach of ~\cite{24} typically detected consecutive blinks as a single blink. This created a problem for subsequent steps of drowsiness detection, since multiple consecutive blinks can be a sign of drowsiness. We added a post-processing step (Blink Retrieval Algorithm), and applied on top of the output of \cite{24}, so as to successfully identify the multiple blinks which may be present in a single detection produced by \cite{24}. Our post-processing step, while lengthy to describe, relies on heuristics and does not constitute a research contribution. To allow our results to be duplicated, we provide the details of that post-processing step as supplementary material.

\begin{figure}  
  \begin{subfigure}[b]{0.243\textwidth}
    \includegraphics[width=\linewidth]{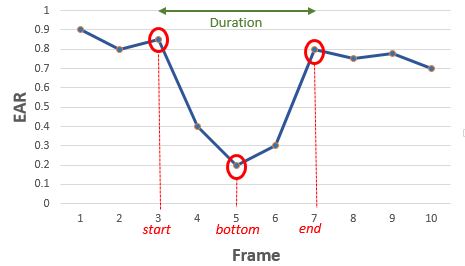}
    \caption{}
    \label{fig:blink}
  \end{subfigure}
  \hfill
  \begin{subfigure}[b]{0.20\textwidth}
    \includegraphics[width=\linewidth]{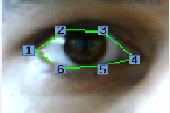}
    \caption{}
    \label{fig:eye}
  \end{subfigure}
  \caption{(a) The EAR sequence during an entire blink and the start, bottom and end points. (b) The eye landmarks to define EAR for each frame.
  }
\end{figure}

The input to the blink detection module is the entire video (with a length of approximately ten minutes in our dataset). In a real-world application of drowsiness detection, where a decision should be made every few minutes, the input could simply consist of the last few minutes of video. The output of the blink detection module is a sequence of blink events $\{\mathbf{blink}_1, ..., \mathbf{blink}_K\}$. Each $\mathbf{blink}_i$ is a four-dimensional vector containing four features describing the blink: duration, amplitude, eye opening velocity, and frequency. For each blink event $\mathbf{blink}_i$, we defined $start_i$, $bottom_i$, and $end_i$ as the ``start'', ``bottom'' and ``end'' points (frames) in that blink (Fig.\ref{fig:blink}) explained in the Blink Retrieval Algorithm. Also, for each frame $k$, we denoted:
\begin{small}
\begin{equation}
EAR[k]=\frac{||\vec{p_2}-\vec{p_6}||+||\vec{p_3}-\vec{p_5}||}{||\vec{p_1}-\vec{p_4}||}
\end{equation}
\end{small}where $\vec{p_i}$ is the 2D location of a facial landmark from the eye region (Fig.\ref{fig:eye}).
Using this notation, we define four main scale invariant features that we extract from $\mathbf{blink}_i$. These are the features that we use for our baseline drowsiness detection method:
\begin{small}
\begin{equation}
\text{Duration}_i= end_i-start_i+1
\end{equation}
\vspace{0.05in}
\begin{equation}
\text{Amplitude}_i= \frac{EAR[start_i]-2 EAR[bottom_i]+EAR[end_i]}{2}
\end{equation}
\begin{equation}
\text{Eye Opening Velocity}_i= \frac{EAR[end_i]-EAR[bottom_i]}{end_i-bottom_i}
\end{equation}
\vspace{0.05in}
\begin{equation}
\text{Frequency}_i= 100\times \frac{\text{Number of blinks up to blink}_i}{\text{Number of frames up to }end_i}
\end{equation}
\end{small}  
\subsection{Drowsiness Detection Pipeline}
\label{subsection_pipeline}
\textbf{Preprocessing:} 
A big challenge in using blink features for drowsiness detection is the difference in blinking pattern across individuals~\cite{4,6,15,soha}, so features should be normalized across subjects if we are going to train the whole data together at once. In order to tackle this challenge, we use the first third of the blinks of the \textbf{alert} state to compute the mean and standard deviation of each feature for each individual, and then use Equation \ref{norm} to normalize the rest of the alert state blinks as well as the blinks in the other two states of the same person($m$) and feature($n$):
\begin{small}
\begin{equation}\label{norm}
\overline{\mathrm{Feature}}_{n,m}= \frac{\mathrm{Feature}_{n,m}-\mu_{n,m}}{\sigma_{n,m}}
\end{equation}
\end{small}
Here, $\mu_{n,m}$ and $\sigma_{n,m}$ are the mean and standard deviation of feature $n$ in the first third of the blinks of the alert state video for subject $m$.

We do this normalization for both the training and test data of all subjects and features. A similar approach has been taken in ~\cite{6,15}. This normalization is a realistic constraint: when a driver starts driving a new car or a worker starts working, the camera can use the first few minutes (during which the person is expected to be alert) to compute the mean and variance, and calibrate the system. This calibration can be used for all subsequent trips or sessions. The detector decides the state of the subject relative to the statistics collected during the calibration stage. 
We should clarify that, in our experiments, the alert state blinks used for normalization are never used again either for training or testing. After the per-individual normalization, we perform a second normalization step, where we normalize each feature so that, across individuals, the distribution of the feature has a mean of zero and a variance of one.

\textbf{Feature Transformation Layer: }Instead of defining a large number of features initially, and then selecting the most relevant ones\cite{4}, we let the network use the four main blink features and learn to map them to a higher dimensional feature space to minimize the loss function. The goal of the fully connected layer before the HM-LSTM module is to take each 4D feature vector at each time step as input and transform it to an $L$ dimensional space with shared weights ($W\in\mathbb{R}^{4\times L}$) and biases ($\mathbf{b}\in\mathbb{R}^{1\times L}$) across time steps. Define $\mathbf{T}$ as the number of time steps used for the HM-LSTM Network and $\mathbf{f}_i\in\mathbb{R}^{1\times L}$ for each blink at each time step $i$ , so that:
\begin{small}
\begin{equation}\label{pre}
F=\text{ReLU}(BW+\overline{b})
\end{equation}
\end{small}where $F=\left[\mathbf{f}^T_1,\mathbf{f}^T_2, ...,\mathbf{f}^{T}_\mathbf{T}\right]^T$,  $\overline{b}=\left[\mathbf{b}^T,\mathbf{b}^T, ...,\mathbf{b}^{T}\right]^T$,\\
$\overline{b}\in\mathbb{R}^{\mathbf{T}\times L}$ and $B=\left[\mathbf{blink}^T_1,\mathbf{blink}^T_2, ...,\mathbf{blink}^T_\mathbf{T}\right]^T$. 

\textbf{HM-LSTM Network: }
Our approach introduces a temporal model to detect drowsiness. The work by\cite{8}, using Hidden Markov Model (HMM), suggests that drowsiness features follow a pattern over time. Thus, we used an HM-LSTM network\cite{23} to leverage the temporal pattern in blinking. 
It is also ambiguous how each blink is related to the other blinks or how many blinks in succession can affect each other. To remedy this challenge, we used HM-LSTM cells to discover the underlying hierarchical structure in a blink sequence.

Chung \etal\cite{23} introduces a parametrized boundary detector, which outputs a binary value, in each layer of a stacked RNN. For this boundary detector, positive output for a layer at a specific time step signifies the end of a segment corresponding to the latent abstraction level for that layer. Each cell state is ``updated'', ``copied'' or ``flushed'' based on the values of the adjacent boundary detectors. As a result, HM-LSTM networks tend to learn fine timescales for low-level layers and coarse timescales for high-level layers. This dynamic hierarchical analysis allows the network to consider blinks both in short and long segments, depending on when the boundary detector is activated for each cell. For additional details about HM-LSTM, we refer the readers to \cite{23}.

The HM-LSTM network takes each row of $F$ as input at each time step and outputs a hidden state $\mathbf{h}_l\in \mathbb{R}^{1\times H}$ only at the last time step for each layer $l$. $H$ is the number of hidden states per layer.

\textbf{Fully Connected Layers: }We added a fully connected layer (with $W_{1,l}\in \mathbb{R}^{H\times L_1}$ as weights and $\mathbf{b}_{1,l}\in \mathbb{R}^{1\times L_1}$ as biases) to the output of each layer $l$ with $L_1$ units to capture the results of the HM-LSTM network from different hierarchical perspectives separately.  Define $\mathbf{e}_{1l}\in \mathbb{R}^{1\times L_1}$ for each layer, so that:
\begin{small}
\begin{equation}
\mathbf{e}_{1l}=\text{ReLU}(\mathbf{h}_l W_{1,l}+\mathbf{b}_{1,l})
\end{equation}
\end{small}
Then, we concatenated $\mathbf{e}_{1 l}$ $\forall$ $l\in\{i|i=1,2, ...,\overline{L}\}$ to form $\mathbf{e}_1=\left[\mathbf{e}_{11},\mathbf{e}_{12}, ...,\mathbf{e}_{1\overline{L}}\right]$, where $\mathbf{e}_1\in \mathbb{R}^{1\times (L_1.\overline{L})}$ and $\overline{L}$ is the number of layers.

Similarly, as shown in Fig. \ref{architecture}, $\mathbf{e}_1$ is fed to more fully connected layers (with ReLU as their activation functions) in FC2,FC3 and FC4, resulting in $\mathbf{e}_4\in \mathbb{R}^{1\times (L_4)}$, where $L_4$ is the number of units in FC4.

\textbf{Regression Unit: }A single node at the end of this network determines the degree of drowsiness by outputting a real number from 0 to 10 depending on how alert or drowsy the input blinks are (Eq.\ref{lone_unit}). This 0 to 10 scale helps the network to model the natural transition from alertness to drowsiness unlike the previous works~\cite{7,9}, where inputs were classified directly into different classes discretely. 
\begin{small}
\begin{equation}\label{lone_unit}
out=10\times\text{Sigmoid}(\mathbf{e}_4 W_o+b_o)
\end{equation}
\end{small}
Here, $W_o\in \mathbb{R}^{L_4\times 1}$ and $b_o\in \mathbb{R}^{1\times 1}$ are the regression parameters, and $out\in \mathbb{R}^{1\times 1}$ is the final regression output.

\textbf{Discretization and Voting: }When someone is drowsy, it does not mean that all their blinks will necessarily represent drowsiness. As a result, it is important to classify the drowsiness level of each video as the most dominant state predicted from all blink sequences in that video. As the first step, we used Eq.\ref{class} to discretize the regression output to each of the predefined classes. 
\begin{small}
\begin{equation}\label{class}
\text{class}(out)=\left\{\begin{array}{cl}
Alert,&0.0\leq out < 3.3\\
Low Vigilant,&3.3\leq out \leq 6.6\\
Drowsy,&6.6< out \leq 10 
\end{array}\right.
\end{equation}
\end{small}
Suppose there are K blinks in video V. Using a sliding window of length T, each T consecutive blinks form a blink sequence that is given as input to the network (Eq.\ref{pre}), resulting in possibly multiple blink sequences. The most frequent predicted class from these multiple sequences would be the final classification result of video V. The positive effect of voting is shown later in our results.

\textbf{Loss Function: } Our model learns not to penalize predictions ($out_i$) that are within a certain distance $\sqrt{\Delta}$ of true labels ($t_i$) for all $N$ training sequences, and instead penalizes less accurate predictions quadratically by their squared error. As a result, our model is more concerned about classifying each sequence correctly rather than perfect regression. This attribute helps us to jointly do regression and classification by minimizing the following loss function:
\begin{small}
\begin{equation}\label{loss}
loss=\frac{\sum_{i=1}^{N} \text{max}(0,\vert{out_i-t_i}\vert^2-\Delta)}{N}
\end{equation}
\end{small}
\section{Experiments}
\subsection{Evaluation Metrics }
We designed four metrics to fully evaluate our model from different views and at various stages of the pipeline.

\textbf{Blink Sequence Accuracy (BSA)}:
This metric evaluates the results before ``the voting stage'' and after ``discretization'' across all test blink sequences. 

\textbf{Blink Sequence Regression Error (BSRE)}:
We define BSRE as follows:
\begin{small}
\begin{equation}\label{BSRE}
BSRE=\frac{\sum_{i=1}^{M}C^s_i\vert{out_i-S_i}\vert^2}{M}
\end{equation}
\end{small}

In the above equation, $C^s_i$ is a binary value, equal to 0 if the $i$-th blink segment has been classified correctly, and equal to 1 otherwise. Eq.\ref{BSRE} penalizes each wrongly classified blink sequence $i$ by a term quadratic to the distance of the regressed output to the nearest true state border ($S_i$) defined in Eq.\ref{class}. Blink sequences classified correctly do not contribute to the BSRE error.

\textbf{Video Accuracy (VA)}:
``Video Accuracy'' is the main metric of accuracy, it is equal to the percentage of entire videos (not individual video segments) that have been classified correctly.

\textbf{Video Regression Error (VRE)}:
VRE is defined as:
\begin{small}
\begin{equation}\label{VRE}
VRE=\frac{\sum_{j=1}^{Q}C^v_j \vert{\frac{1}{K_j}\sum_{i=1}^{K_j} (out_{i,j})-S_j}\vert^2}{Q}
\end{equation}
\end{small}
In the above, Q is the total number of videos in the test set, and $C^v_j$ is a binary value, equal to 0 if the $j$-th video has been classified correctly and equal to 1 otherwise. $K_j$ is the number of all blink sequences in video $j$. Correctly classified videos do not contribute at all to the VRE error. For a fixed VA, the value of VRE indicates the margin of error for wrongly classfied videos.

\begin{figure}  
  \begin{subfigure}[b]{0.22\textwidth}
    \includegraphics[width=\linewidth]{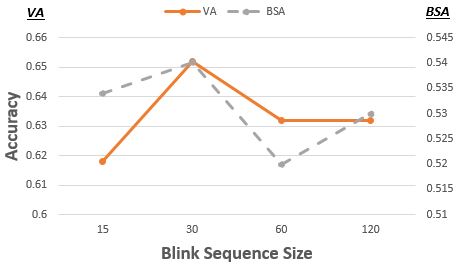}
    \caption{}
    \label{size}
  \end{subfigure}
  \hfill
  \begin{subfigure}[b]{0.22\textwidth}
    \includegraphics[width=\linewidth]{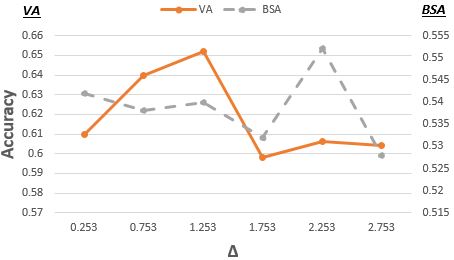}
    \caption{}
    \label{delta}
  \end{subfigure}
  \caption{The effect of blink sequence size and $\Delta$ to the accuracy metrics.
  }
\end{figure}
\subsection{Implementation }
We used one fold of the RLDD dataset as our test set, and the remaining four folds for training. After repeating this process for each fold, the results were averaged across the five folds. For parameter $\mathbf{T}$ defined in Section \ref{subsection_pipeline}, which specifies the number of consecutive blinks provided as input to the network, we used a value of 30 (Fig. \ref{size}). Videos with less than 30 blinks were zero padded. Blink sequences were generated by applying this sliding window of 30 blinks on each video, with a stride of two. If the window size is too large, the long dependency on previous blinks can significantly delay the correct output while transitioning from one state to the other.  

We annotated all sequences with the label of the video they were taken from. Our model was trained on around 7000 blink sequences (depending on the training fold) using Adam optimizer~\cite{adam} with a learning rate of 0.000053, $\Delta$ of 1.253 (Fig.\ref{delta}), and batch size of 64 for 80 epochs in all five folds. We also used batch normalization and L2 regularization with a coefficient ($\lambda$) of 0.1. The HM-LSTM module has four layers with 32 hidden states for each layer. More details about the architecture is shown in Fig.\ref{architecture}. 
\subsection{Experimental Results}\label{humanexp}
In this section, we evaluate our baseline method with respect to the human judgment benchmark explained in section \ref{human}. Due to lack of a state-of-the-art method on a realistic and public dataset, we compare our baseline method with two variations of our pipeline to show that the whole pipeline performs best with HM-LSTM cells. The first version has the same architecture, as our network, with typical LSTM cells~\cite{LSTM} used instead of HM-LSTM cells. The second version is a simpler version with the same architecture after removing the HM-LSTM module, where the input sequence is fed to a fully connected multilayer network.
\begin{table}
\begin{center}
\footnotesize\setlength{\tabcolsep}{2.5pt}
\begin{tabular}{l@{\hspace{2pt}} *{4}{c}}
\toprule
\bfseries Model & \multicolumn{4}{c}{\bfseries Evaluation Metric} \\
\cmidrule(l){2-5}
& BSRE &VRE & BSA & VA  \\
\midrule
\rowcolor[HTML]{C0C0C0} 
\textit{HM-LSTM network}
& \textbf{1.90} & \textbf{1.14} & \textbf{54\%} & \textbf{65.2\%} \\
\textit{LSTM network}
& 3.42 & 2.68 & 52.8\% & 61.4\% \\
\rowcolor[HTML]{C0C0C0} 
\textit{Fully connected layers}
& 2.85 & 2.17 & 52\% & 57\% \\
\textit{Human judgment}
& | & 2.01 & | & 57.8\% \\
\bottomrule
\end{tabular}
\end{center}
\caption{This table numerically compares the performance of our model with two simplified versions of the network and human judgment using four predefined metrics. The above values are the final averaged values across all test folds. }\label{tab:full}
\end{table}

The results of our comparison with these two variations and the human judgment benchmark are listed in Table \ref{tab:full}. This table shows the final cross validation results of drowsiness detection by the predefined metrics.
This comparison not only highlights the temporal information in blinks, but also shows the 4\% increase in accuracy we gained after switching to HM-LSTM from typical LSTM cells. As indicated by BSRE and VRE metrics in Table \ref{tab:full}, the margin of error for regression is also considerably lower in the HM-LSTM network compared to the other two.
The results for LSTM and HM-LSTM networks suggest that temporal models provide better solutions for drowsiness detection than simple fully connected layers.

As mentioned before, all blink sequences in each video were labeled the same. However, in reality, not all blinks represent the same level of drowsiness. This discrepancy is an important reason that BSA is not high, and ``voting'' makes up for that resulting in a higher accuracy in VA.

\begin{figure}  
  \begin{subfigure}[b]{0.235\textwidth}
    \includegraphics[width=\linewidth]{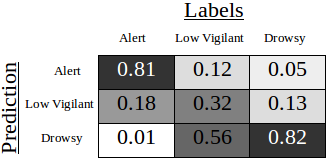}
    \caption{}
    \label{confusion}
  \end{subfigure}
  \hfill
  \begin{subfigure}[b]{0.235\textwidth}
    \includegraphics[width=\linewidth]{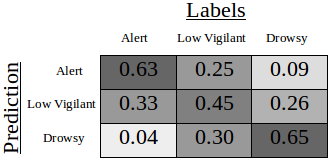}
    \caption{}
    \label{confusion2}
  \end{subfigure}
  \caption{Confusion matrices for: (a) our proposed model  and (b) human judgment results (video accuracy).}
\end{figure}

Fig.\ref{confusion} shows that the middle class (low vigilant) is, as expected, the hardest to classify, where it is mostly misclassified for ``drowsy''. On the other hand, our model classifies alert and drowsy subjects very confidently with over 80\% accuracy, and rarely misclassifies alertness for drowsiness or vice versa. This means, that the results are mostly reliable in practice.

In addition, our model detects early signs and subtle cases of drowsiness better than humans in the RLDD dataset by just analyzing the temporal blinking behavior. The detailed quantitative results for all folds and the final averaged values are listed in Table \ref{tab:human} and Table \ref{tab:full} respectively.
Our drowsiness detection model has approximately 50,000 trainable parameters. Storing those parameters does not occupy much memory space, and thus the model can be easily stored on even low-end cell phones. In terms of running time (at the evaluation stage, after training ), the end-to-end system  processes approximately 35-80 frames per second ( for the frame size range of 568x320 to 1920x1080), on a Linux workstation with an Intel Xeon CPU E3-1270 V2 processor running at 3.5 GHz, and with 16GB of memory.
\begin{table}
\begin{center}
\footnotesize\setlength{\tabcolsep}{3.5pt}
\begin{tabular}{l@{\hspace{2pt}} *{10}{c}}
\toprule
\bfseries Case & \multicolumn{10}{c}{\bfseries Metric-Fold} \\
\cmidrule(l){2-11}
& A-f1 & R-f1 & A-f2 & R-f2 & A-f3 & R-f3 & A-f4 & R-f4 & A-f5 & R-f5 \\
\midrule
\bfseries PM 
& 0.64 & 2.42 & 0.61 & 1.04 & 0.70 & 0.58 & 0.64 & 0.85 & 0.67 & 0.81\\
\bfseries HJ
& 0.62 & 1.37 & 0.59 & 2.3 & 0.60 & 1.96 & 0.53 & 2.32 & 0.55 & 2.07 \\
\bottomrule
\addlinespace
\multicolumn{11}{l}{A-f i: VA for fold i}\\
\multicolumn{11}{l}{R-f i: VRE for fold i}
\end{tabular}
\end{center}
\caption{Results of our Proposed Model (PM) and Human Judgment (HJ) measured by VA and VRE}\label{tab:human}
\end{table}
\section{Conclusions}
In this paper, we presented a new and publicly available real-life drowsiness dataset (RLDD), which, to the best of our knowledge, is significantly larger than existing datasets, with almost 30 hours of video. We have also proposed an end-to-end baseline method using the temporal relationship between blinks for multistage drowsiness detection. The proposed method has low computational and storage demands. Our results demonstrated that our method outperforms human judgment in two designed metrics on the RLDD dataset.

One possible topic for future work is to add a spatial deep network to learn other features of drowsiness besides blinks in the video. In general, moving from handcrafted features to an end-to-end learning system is an interesting topic, but the amount of training data that would be necessary is not clear at this point. Overall, we hope that the proposed public dataset will also encourage other researchers to work on drowsiness detection and produce additional and improved results, that can be duplicated and compared to each other.

\section*{Acknowledgement}

This work was partially supported by National Science Foundation
grants IIS 1565328 and IIP 1719031.

\clearpage

{\small
\bibliographystyle{ieee}
\bibliography{egbib}
}
\clearpage

\title{Supplementary Material}

\maketitle
\thispagestyle{empty}

%%%%%%%%% BODY TEXT
\section*{Blink Retrieval Algorithm}

% beginning of stuff that should go to supplementary material
In our experiments, we noticed that the approach of  Soukupov\'{a} and Cech~\cite{24} typically detected consecutive quick blinks as a single blink. This created a problem for subsequent steps of drowsiness detection, since multiple consecutive blinks can typically be a sign of drowsiness. We added a post-processing step on top of the output of \cite{24}, that successfully identifies the multiple blinks which may be present in a single detection produced by \cite{24}.

According to\cite{24}, define EAR, for each frame, as below:
\begin{small}
\begin{equation}
EAR=\frac{||\vec{p_2}-\vec{p_6}||+||\vec{p_3}-\vec{p_5}||}{||\vec{p_1}-\vec{p_4}||}
\end{equation}
\end{small}

\begin{figure}[t]
\begin{center}

   \includegraphics[width=0.4\linewidth]{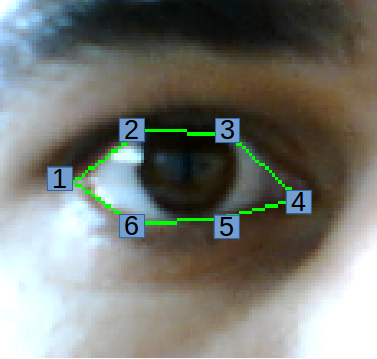}
\end{center}
   \caption{Six points marking each eye.}
\label{eye}
\end{figure}

\begin{figure}[t]
\begin{center}

   \includegraphics[width=1.0\linewidth]{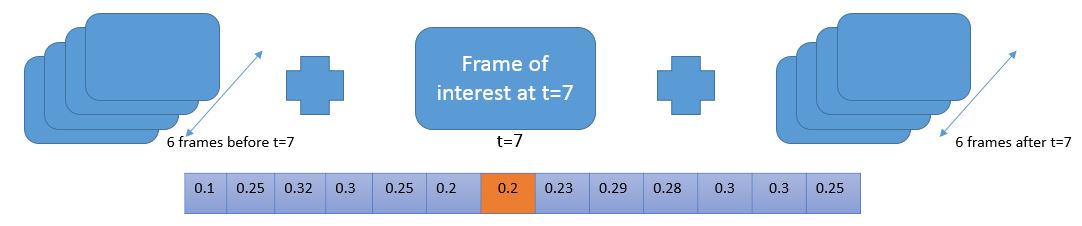}
\end{center}
   \caption{Presenting each frame (at t=7) by 13 numbers (EARs) concatenated from 13 frames as a feature vector.}
\label{13D}
\end{figure}

In the above, each $\vec{p_i}\in\{p_i|i=1,...,6\}$ is the 2D location of a facial landmark from the eye region, as illustrated by Figure \ref{eye}. In~\cite{24}, an SVM classifier detects eye blinks as a pattern of EAR values in a short temporal window of size 13 depicted in Fig.\ref{13D}. This fixed window size is chosen based on the rationale that each blink is about 13 frames long. A single blink takes around 200ms to 400ms on average \cite{soha,singh}, which translates to six to twelve frames for a video recorded at 30fps. Even if 13 frames is a good estimate for the length of a blink, this approach would not handle consecutive quick blinks.

\begin{figure}
  \begin{subfigure}[b]{0.4\textwidth}
    \includegraphics[width=\linewidth]{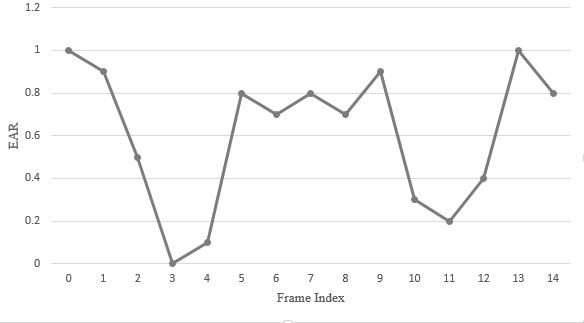}
    \caption{}
    \label{fig:plain_x}
  \end{subfigure}
  \hfill
  \begin{subfigure}[b]{0.4\textwidth}
    \includegraphics[width=\linewidth]{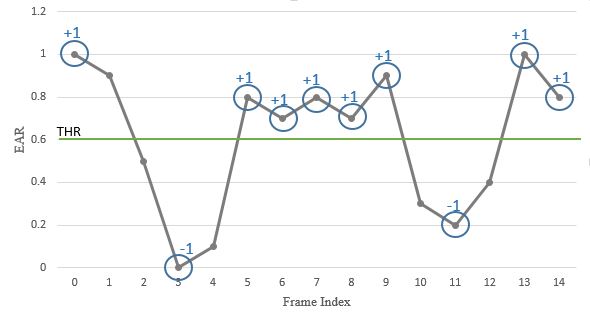}
    \caption{}
    \label{fig: extrema}
  \end{subfigure}

  \begin{subfigure}[b]{0.4\textwidth}
    \includegraphics[width=\linewidth]{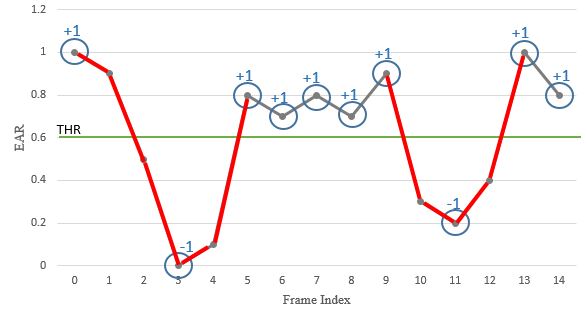}
    \caption{}
    \label{fig:legs}
  \end{subfigure}
  \hfill
  \begin{subfigure}[b]{0.4\textwidth}
    \includegraphics[width=\linewidth]{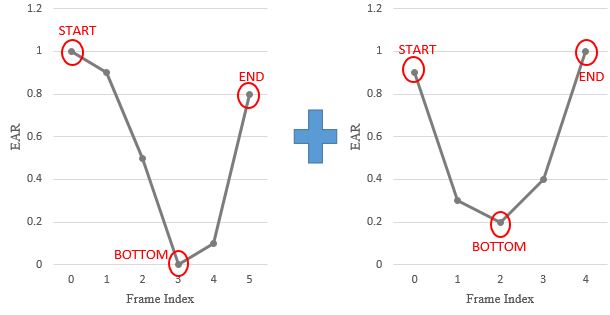}
    \caption{}
    \label{blinks}
  \end{subfigure}
  \caption{The Blink Retrieval Algorithm steps: (a) $\mathbf{x}$ with size $M=15$ as the input for Alg. \ref{alg:Blink Retrieval Algorithm}. (b) The indices of circled points form $\mathbf{e}$, and the set of +1 and -1 labels forms $\mathbf{t}$ with $P=8$. (c) The red lines indicate where $\mathbf{z}$ values are negative. (d) Two ($N=2$) blinks are retrieved with definite start, end and bottom points.
  }
\end{figure}

As depicted in Figure \ref{13D}, each value in this 13 dimensional vector corresponds to the EAR of a frame with the frame of interest located in the middle. The SVM classifier takes these 13D vectors as input and classifies them as ``open'' or ``closed'' (more specifically referred to the frame of interest in each input vector). A number of consecutive ``closed'' labels represent a blink with the length of $M$. Subsequently, the EAR values of these $M$ frames are stored in $\mathbf{x}$ in order, and fed to the ``Blink Retrieval Algorithm'', explained in Alg.\ref{alg:Blink Retrieval Algorithm}, for post-processing (Fig. \ref{fig:plain_x}). The sequence of EAR values for \textbf{one} blink by~\cite{24} will be considered as a \textbf{candidate} for \textbf{one or more than one} blinks.

This algorithm runs in $\Theta(M)$ time, where $M$ is the number of frames in the video segment that is used as input to the algorithm. In practice, the algorithm runs in real time. In addition,  Alg.\ref{alg:Blink Retrieval Algorithm} sets a definite frame on when a blink starts, ends or reaches its bottom point based on the extrema of its EAR signal. For better results,
$x$ is passed through a median/mean filter to clear the noise and then fed to the algorithm.

At step 1, the derivative of $\mathbf{x}$ is taken. Then, zero derivatives are modified, at steps 2 and 3, so that those derivatives have the same sign as the derivative at their previous time step. This modification helps to find local extrema, as points where the derivative sign changes (steps 4 to 7). The threshold, defined at step 8, is used to suppress the subtle ups and downs in $\mathbf{x}$ due to noise and not blinks. The extrema in $\mathbf{x}$ are circled in Figure \ref{fig: extrema}, and labeled (+1 or -1) relative to the threshold (steps 9 to 11). Each two consecutive extrema are indicative of a downward or upward movement of eyes in a blink if those two are connected, so that the link or links between them pass the threshold line (steps 12 and 13). Fig.\ref{fig:legs} highlights these links in red. Finally, each pairing of these red links corresponds to one blink with start, end and bottom points as depicted in Figure \ref{blinks} (steps 14 to the end).\\

%-------------------------------------------------------------------------

\begin{algorithm}
\caption{Blink Retrieval Algorithm}\label{alg:Blink Retrieval Algorithm}
\hspace*{\algorithmicindent} \textbf{Input}{ The initial detected EAR signal $\mathbf{x}\in \mathbb{R} ^M$, where $M$ is the size of the $\mathbf{x}$ time series, as a candidate for one or more blinks and epsilon=0.01} \\
\hspace*{\algorithmicindent} \textbf{Output} {$N$ retrieved blinks, $N\ll M$}
\begin{algorithmic}[1]
\State $\mathbf{\dot{x}}[n]\gets \mathbf{x}[n+1]-\mathbf{x}[n],$ $\forall$ $n\in\{i| i=0,1,...,M-2\}$
\If {$\mathbf{\dot{x}}[0]=0$} 
$\mathbf{\dot{x}}[0] \gets -1\times\text{epsilon}$
\EndIf 
\State $\mathbf{\dot{x}}[n] \gets \mathbf{\dot{x}}[n-1]\times\text{epsilon,}$ $\forall$ $n\in\{i|\mathbf{\dot{x}}[i]=0 \land i\ne 0\}$ to avoid zero derivatives for steps 4 and 6
\State $\mathbf{c}[n] \gets \mathbf{\dot{x}}[n+1]\times\mathbf{\dot{x}}[n],$ $\forall$ $n\in\{i| i=0,1,...,M-3\}$
\State Define $\mathbf{e}\in \mathbb{R}^{P+2}, P\le M-2$  to store the indices for the $P$ extrema, the first and the last points in $\mathbf{x}$ 
\State $\mathbf{e}[0]\gets 0$, $\mathbf{e}[P+1]\gets M-1$, supposing the first and last points in $\mathbf{x}$ are maxima
\State $\mathbf{e}[k]\gets n+1,$ $\forall$ $(n\in\{i|\mathbf{c}[i]<0 \} \land k\in\{i|i=1,2,...,P\})$  \Comment{Indices of $P$+2 extrema, including the first and last points in $\mathbf{x}$ are stored in order}
\State $\text{Define } THR\gets 0.6\times\max(\mathbf{x})+0.4\times\min(\mathbf{x})$, as a threshold
\State $\text{Define } \mathbf{t}\in \mathbb{R}^{P+2}$
 , to store +1 or -1 for extrema above and below threshold respectively
\State $\mathbf{t}[0]\gets +1, \mathbf{t}[P+1]\gets +1$, supposing the first and last points in $\mathbf{x}$ are maxima
\State Append +1 in $\mathbf{t}$ for each $n\in\{i| \mathbf{x}[\mathbf{e}[i]]>THR\}$, and append -1 in $\mathbf{t}$ for each $n\in\{i| \mathbf{x}[\mathbf{e}[i]]\le THR\}$, all in the order of the indices in $\mathbf{e}$
\State Define $\mathbf{z}\in \mathbb{R}^{P+1},$ $\mathbf{z}[n]\gets \mathbf{t}[n+1]\times\mathbf{t}[n]$
\State Define $\mathbf{s}$, to store the indices of all negative values in $\mathbf{z}$, representing the downward and upward movements of eyes in a blink 
\State $N\gets \frac{\textit{length}(\mathbf{s})}{2}$ \Comment{$N$ is the number of sub blinks, and   $\textit{length}(\mathbf{s})$ is always an even number}
\For{$i\gets $0 to ${N-1}$}  \Comment{Define for $blink_i$:}
	\State $\text{   }$ StartIndex $\gets \mathbf{e}[\mathbf{s}[2\times i]] $,
	\State $\text{   }$ EndIndex$\gets \mathbf{e}[\mathbf{s}[2\times i+1]+1]$,
	\State $\text{   }$ BottomIndex$\gets \mathbf{e}[\mathbf{s}[2\times i+1]]$
\EndFor
\Return start, end and bottom points of the $N$ retrieved blinks in $\mathbf{x}$
\end{algorithmic}
\end{algorithm}

\clearpage

\end{document}